\pdfoutput=1

\documentclass[11pt]{article}
\usepackage{hyperref}

\usepackage[preprint]{acl}

\usepackage{times}
\usepackage{latexsym}
\usepackage[most]{tcolorbox}
\usepackage{subcaption}
\usepackage{booktabs}
\usepackage{multirow} 
\usepackage{array}
\usepackage{amssymb}
\usepackage{balance}

\usepackage[T1]{fontenc}

\usepackage[utf8]{inputenc}

\usepackage{microtype}

\usepackage{inconsolata}

\usepackage{graphicx}
\graphicspath{{./figures/}}

\title{ELTEX: A Framework for Domain-Driven Synthetic Data Generation}

\author{
  \textbf{Arina Razmyslovich\textsuperscript{1}},
  \textbf{Kseniia Murasheva\textsuperscript{1}},
  \textbf{Sofia Sedlova\textsuperscript{2}},
\\
  \textbf{Julien Capitaine\textsuperscript{2}},
  \textbf{Eugene Dmitriev\textsuperscript{1}}
\\
\\
  \textsuperscript{1}Distributed Networks Institute (DNI),
  \textsuperscript{2} Technologies Mésozoïques
\\
\\
  \small{
    \textbf{Correspondence:} \href{mailto:arina.razmyslovich@dn.institute}{arina.razmyslovich@dn.institute}
  }
}

\begin{document}
\maketitle
\begin{abstract}
We introduce Efficient LLM Token Extraction (ELTEX), a framework addressing the critical challenge of LLM domain specialization by systematically extracting and integrating domain indicators throughout synthetic data generation. Unlike approaches relying on implicit knowledge transfer, ELTEX explicitly leverages domain signals to maintain specialized knowledge integrity. In our cybersecurity case study, ELTEX-enhanced data enables a fine-tuned Gemma-2B model to achieve performance competitive with GPT-4o on blockchain cyberattack classification while reducing computational requirements. Our Google Sheets implementation makes ELTEX accessible to non-technical users. Our contributions include: (1) the ELTEX framework; (2) Google Sheets Add-on implementation; (3) empirical validation showing how ELTEX bridges performance gaps between small and large models; and (4) a synthetic dataset of 11,448 texts for blockchain cyberattack detection.
\end{abstract}

\section{Introduction}
Domain specialization remains a critical challenge for deploying Large Language Models (LLMs) in industry settings \cite{ling2024domainspecializationkeymake}. While LLMs demonstrate impressive general capabilities, their performance deteriorates when confronted with specialized terminology and contextual nuances found in professional domains \cite{kim2024bridging, lu2024fine}.

We introduce Efficient LLM Token Extraction (ELTEX), a domain-driven framework that systematically integrates explicit domain indicator extraction with dynamic prompting to preserve critical domain nuances throughout the synthetic data generation process. ELTEX addresses a fundamental limitation in existing approaches: the difficulty of maintaining domain-specific knowledge integrity when generating training data for specialized applications. Unlike general-purpose data generation, ELTEX explicitly identifies and leverages domain indicators to guide the generation process.

Existing approaches to synthetic data generation have primarily relied on implicit knowledge transfer through few-shot examples or post-generation validation \cite{patil2024review}. Recent advances have introduced more sophisticated methodologies: UniGen \cite{wu2024unigenunifiedframeworktextual} implements attribute-guided generation with RAG-based validation, while CRAFT \cite{ziegler2024craft} employs corpus-based retrieval with few-shot learning. However, these methods still face challenges in explicitly extracting and preserving domain-specific knowledge throughout the generation pipeline.

The effectiveness of ELTEX is particularly evident in cybersecurity applications, where benchmarks such as CyberBench \cite{liu2024cyberbench} reveal that even state-of-the-art LLMs (e.g., GPT-4) struggle with domain-specific tasks. For example, in our case study, we demonstrate ELTEX's effectiveness through text classification of blockchain-related cyberattack discussions, enabling a fine-tuned Gemma-2B model \cite{gemmateam2024gemmaopenmodelsbased} to achieve performance competitive with GPT-4o while requiring significantly fewer computational resources.

To enhance accessibility, we have deployed ELTEX as a Google Sheets add-on, making synthetic data generation accessible to non-technical users. This deployment has significantly reduced the time required to create specialized training datasets, enabling rapid response to emerging domain-specific patterns across various industries.

Our contributions include: (i) ELTEX, a framework that systematically identifies and leverages domain indicators to generate high-fidelity, context-grounded synthetic data; (ii) a Google Sheets add-on enabling non-technical users to generate and manage datasets; \footnote{\href{https://github.com/Kseymur/eltex-sheets-addon}{ELTEX Add-on}} (iii) empirical validation showing that ELTEX-generated data bridges the performance gap between small and large models in specialized tasks; and (iv) a carefully curated synthetic dataset of 11,448 social media texts for cyberattack detection in blockchain. \footnote{The dataset is available on \href{https://huggingface.co/datasets/dn-institute/cyberattack-blockchain-synth}{HuggingFace}}

\section{Related Work}
Specialized domains requiring LLM adaptation often share three critical barriers: (i) privacy constraints that limit data sharing and collection \cite{subramani2023detecting}, (ii) rapid domain drift as terminology, threats, and regulations continuously evolve \cite {lu2018learning}, and thus (iii) insufficient labeled data volume to capture the complexity and nuance of domain-specific tasks. These obstacles create a significant gap between LLMs' general capabilities and their effectiveness in specialized contexts.
 
Recent research tackles these barriers through two complementary approaches: (i) high-fidelity synthetic data generation to alleviate domain-specific data scarcity \cite{ferrag2024generative}, and (ii) compact model architectures (e.g., Microsoft's phi-3-mini \cite{abdin2024phi}), which can be deployed locally to reduce inference costs, preserve privacy and increase throughput.

LLM-based synthetic data generation offers a promising workaround by producing privacy-preserving samples rich in domain-specific details. Conventional approaches rely on heuristic prompts that combine task specifications, generation conditions, and in-context examples to drive LLM output \cite{long-etal-2024-llms}. For more intricate datasets, multi-step generation techniques decompose the process into manageable subtasks. 

Frameworks, such as UniGen \cite{wu2024unigenunifiedframeworktextual} and CRAFT \cite{ziegler2024craft}, have aimed to streamline synthetic data generation, though each faces distinct challenges in knowledge-intensive domains. UniGen provides a unified approach to generating diverse datasets. CRAFT leverages corpus retrieval and LLM augmentation to synthesize task-specific data from few-shot examples, demonstrating success in QA and summarization tasks. However, these frameworks face limitations in knowledge-intensive domains: UniGen struggles with specialized knowledge retention, and CRAFT's reliance on public web corpora limits its efficacy where domain-specific data is scarce and sensitive. Our proposed ELTEX framework builds on these foundations and bridges the gap by systematically extracting and integrating explicit domain knowledge, particularly benefiting knowledge-intensive applications where existing approaches fall short.

\section{ELTEX Framework}
\label{eltexoverview}
\begin{figure}[htbp]
    \centering
    \includegraphics[width=0.48\textwidth]{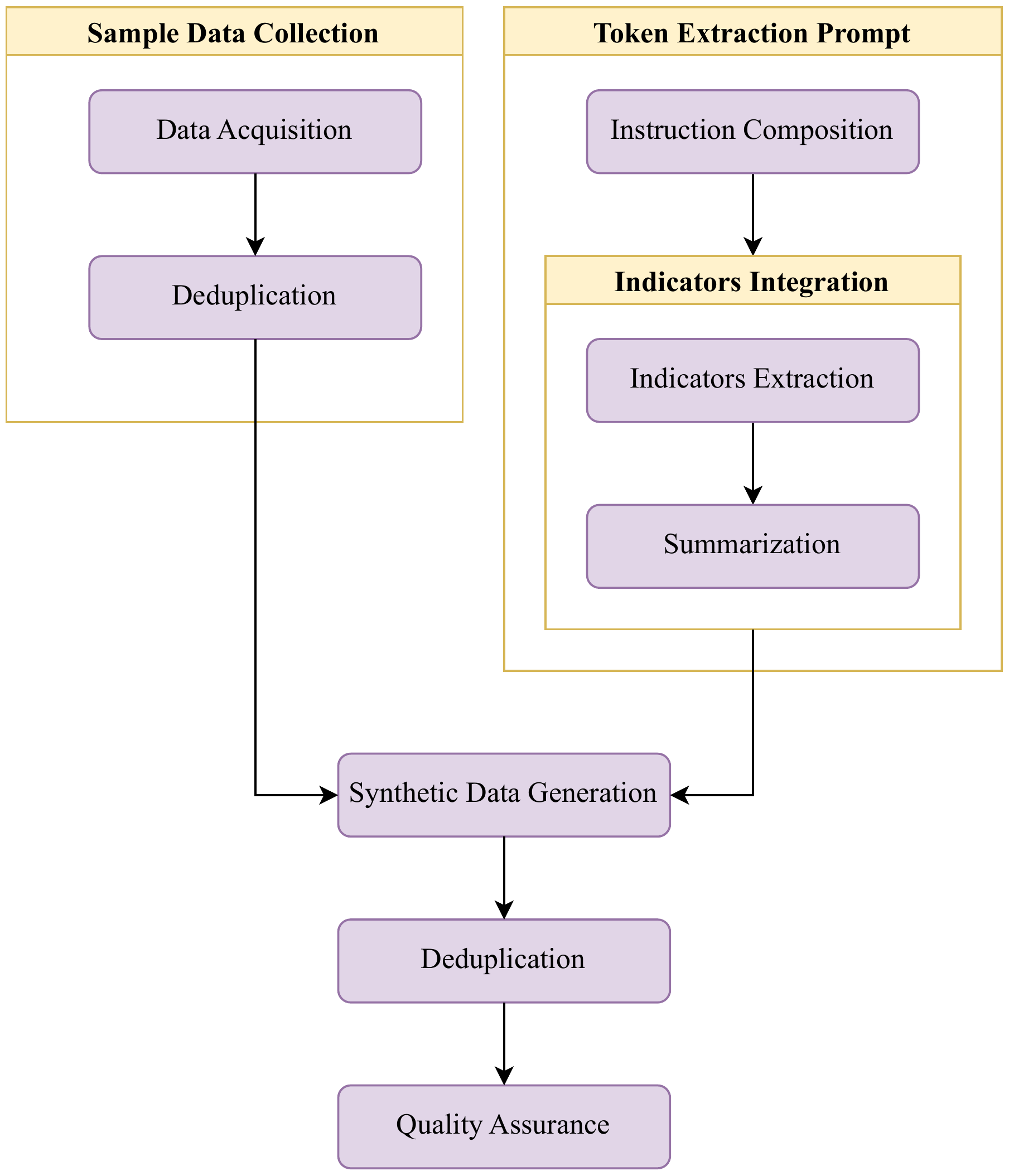}
    \caption{ELTEX Pipeline}
    \label{fig:eltex}
\end{figure}

ELTEX is designed to address domain specialization challenges across industries. As a case study, we demonstrate its application to cybersecurity for detecting blockchain-related cyberattacks in social media. The framework consists of five components (Figure~\ref{fig:eltex}): (1) data collection, (2) token extraction prompt construction, (3) synthetic data generation, (4) deduplication, and (5) post-generation quality assurance (QA). Below, we detail each step using cybersecurity as an illustrative example while emphasizing design choices that ensure adaptability to other domains.

\subsection{Data Collection}
\label{sec:data_collection}
ELTEX requires seed data to guide generation, with collection methods adaptable to industry context. Organizations can leverage: (1) curated seed examples, (2) or synthetic bootstrapping, when even seed examples are scarce, ELTEX can bootstrap from descriptive prompts alone. 

For the case study, we used the official X (formerly Twitter) API (v2)~\cite{x_api} to collect data on cyberattacks from the past three years within the blockchain ecosystem. We retrieved 1,766 initial samples (see Table~\ref{tab:dataset_stats} for statistics). These messages underwent a two-stage annotation process: (1) preliminary annotation using GPT-4o and (2) human review to refine labels. The full annotation process details are in Appendix~\ref{appendix:data_annotation}. All data collection procedures followed ethical guidelines (see Section~\ref{sec:ethical}).

For deduplication, we embedded each message using \texttt{BGE-base-en-v1.5} \cite{bge_embedding} and removed samples exceeding a 0.9 vector similarity threshold. We selected this threshold based on pilot experiments indicating that lower values (e.g., 0.8) eliminated too many valid variations (detailed in Appendix~\ref{appendix:additional_experiments}). 

\subsection{Token Extraction Prompt Construction}
The core innovation of ELTEX is its systematic approach to domain knowledge preservation through explicit token extraction. Unlike approaches relying solely on implicit knowledge transfer via few-shot examples, ELTEX actively identifies and incorporates domain-specific indicators into the generation process. The token extraction process involves querying multiple LLMs to generate candidate domain indicators, followed by consolidation to merge semantically similar items and remove redundancies. This multi-model approach ensures comprehensive coverage, as each LLM may introduce overlooked signals.

We designed a specialized extraction prompt to effectively guide GPT-4o in generating domain-relevant tokens for early cyberattack detection. The prompt integrates instructions on the expected output format, curated lists of potential cyberattack indicators and examples of social media messages. 

For the case study, we used GPT-4o, Claude 3.5~\cite{claude2023}, and Gemini 1.5~\cite{team2024gemini} to generate candidate indicators (e.g., unusual spikes in transaction volume, abnormal confirmation times, unexpected changes in mining difficulty). A detailed example of the constructed prompt, along with a broader discussion of indicator generation and multi-LLM fusion, is provided in Appendix~\ref{appendix:eltex_system}.

\subsection{Synthetic Data Generation}
\label{sec:synthetic_data_generation}
This process is designed to be model-agnostic and compatible with any LLM offering high-quality generation capabilities. The generation approach includes shuffling and batching seed data samples, pairing each batch with the ELTEX prompt containing extracted domain tokens.

Temperature settings moderately influence the diversity and format of synthetic data. Lower temperatures (0.0–0.6) can yield repetitive outputs, medium temperatures (0.7–0.8) balance diversity with domain consistency, and high temperatures (0.9–1.0) often produce outputs that deviate from the desired format. Therefore, we selected a temperature of 0.8 to retain a sufficiently large and diverse dataset. Detailed results, including retention rates at different temperature settings, are provided in Appendix~\ref{appendix:additional_experiments}.

For the case study, we employed GPT-4o via the Azure OpenAI API~\cite{microsoft2025openai}. We shuffled and batched real data samples into groups of 10 messages, paired each batch with the ELTEX prompt, and configured the model to generate 100 synthetic messages per request with a temperature of 0.8. To ensure efficient downstream processing, we utilized structured output formatting~\cite{openai2025structured}, receiving results in a JSON array. Due to the stochastic nature of LLM generation, the actual number of messages per request may vary. This process yielded approximately 16,030 initial synthetic samples. All newly generated messages underwent the same deduplication step described in \ref{sec:data_collection}. 
A cost analysis of the GPT-4o-based synthetic data generation pipeline, including token usage, is provided in Appendix~\ref{appendix:cost_analysis}.

\subsection{Post-generation QA}
While our Google Sheets add-on lacks an integrated QA module, we used NotebookLM \cite{notebooklm} for this specific case study as a non-scalable but effective solution to eliminate misclassified messages. NotebookLM functioned as a general "outlier detector," identifying messages that deviated from their assigned topics. We uploaded our generated data, flagged irrelevant content, and iteratively removed problematic entries until achieving the desired quality. The final dataset contained 11,448 entries (Table~\ref{tab:dataset_stats}). Future ELTEX implementations will integrate dedicated QA capabilities directly into the add-on for greater scalability and efficiency.

\begin{table}[htbp]
    \centering
    \renewcommand{\arraystretch}{1.2} 
    \small 
    \begin{tabular}{|l|r|r|r|}
        \hline
        \textbf{Dataset} & \textbf{Cyberattack} & \textbf{General} & \textbf{Total} \\ \hline
        Real (Initial) & 1,078 & 688 & 1,766 \\ 
        Real (Dedup) & 951 & 652 & 1,603 \\ \hline
        Synthetic (Initial)* & 9,510 & 6,520 & 16,030 \\ 
        Synthetic (Dedup) & 6,941 & 4,524 & 11,465 \\
        Synthetic (Final) & 6,941 & 4,507 & 11,448 \\ \hline
    \end{tabular}
    \caption{Datasets statistics throughout the pipeline. *Initial synthetic count is approximate due to the stochastic nature of LLM generation.}
    \label{tab:dataset_stats}
\end{table}

\section{Google Sheets Add-on}

We specifically designed a Google Sheets add-on (using Apps Script\footnote{\url{https://developers.google.com/apps-script}}) to make the entire ELTEX pipeline accessible to non-technical users such as students or domain experts with minimal programming experience. This tool is already in use within our organization to create real-world datasets to meet internal needs.

Unlike Python-based packages — which require installing libraries, maintaining a local environment, and at least minimal scripting skills — or specialized online sandboxes (e.g., Hugging Face Spaces) that may feel unfamiliar to end-users, our add-on leverages a familiar and widely adopted platform: Google Sheets. 
This design choice reduces deployment overhead (no local installations needed), centralizes data management (Sheets inherently tracks revisions and supports collaboration), and offers a straightforward UI that many domain experts already use in their daily workflows.

\begin{figure*}[ht]
    \centering
    \includegraphics[width=\textwidth]{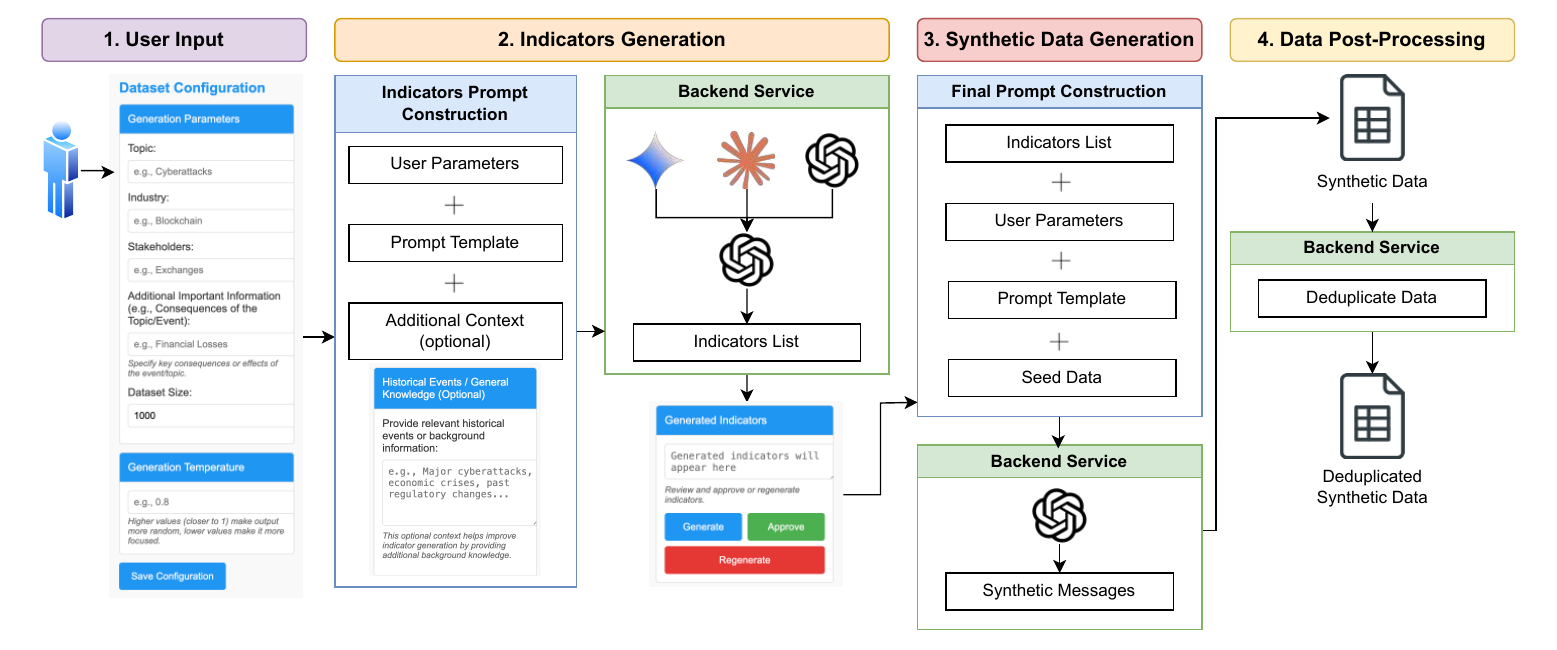}
    \caption{Google Sheets Add-on Workflow. The logos of specific LLM providers are included here as examples only; any similarly capable model could be substituted for each of the depicted LLM services.}
    \label{fig:sheets_diagram}
\end{figure*}

Figure~\ref{fig:sheets_diagram} provides a high-level overview of the system. Below, we summarize the key steps:

\begin{enumerate}
    \item \textbf{User Parameters Input.}
    The user opens a custom sidebar in Google Sheets and specifies the desired parameters for synthetic data generation: 
    \emph{(a)} topic (e.g., “cyberattacks”), 
    \emph{(b)} industry details (e.g., “blockchain”), 
    \emph{(c)} stakeholders (e.g., “exchanges”), 
    \emph{(d)} the dataset size, and 
    \emph{(e)} the generation temperature (defaulting to 0.8 based on our experiments, but fully adjustable by the user).
    \emph{Note:} Because our pipeline may remove duplicates or near-duplicates in subsequent steps, the final retained sample count can be smaller than the user’s requested number.
    
    While some users provide \emph{seed data} (e.g., a handful of real social media messages) for more precise context, others may enter textual descriptions of what they need, allowing the system to generate its initial examples.

    \textit{Additional Context (Optional).}  
    In addition to these parameters, users can supply textual fields labeled \emph{“Historical Events”} or \emph{“General Knowledge”} to guide indicator generation. When querying the LLMs for relevant indicators, the add-on automatically includes this optional background.

    \textit{Advanced Settings.} 
    By default, we select recommended up-to-date LLM providers and models (e.g., GPT-4o, Claude~3.5), but users can override these in the “Advanced Settings” panel, enabling them to choose or update providers/models according to their specific requirements.

    \item \textbf{Indicators Generation.}
    Using the user parameters (and any optional historical context), the tool queries multiple LLMs (e.g., GPT-4o, Claude~3.5) to generate candidate domain indicators. 
    We typically run these indicator-generation calls via a separate microservice to avoid Apps Script’s runtime limits.
    A final summarization step (using a frontier LLM) merges and refines these candidates into one concise list.

    \item \textbf{Combining Seed Data.}
    If the user has provided seed data in the Sheet, the add-on includes them alongside the newly generated indicators and user-defined parameters to form the final prompt. 
    If no seed data is available, the user’s textual descriptions and generated indicators alone guide the synthetic data generation.

    \item \textbf{Synthetic Data Generation.}
    Rather than invoking LLM APIs directly from Apps Script — which has strict execution time limits — we dispatch requests to a separate backend (Cloudflare Workers \cite{Cloudflare}) that handles the actual model calls. 
    Our backend uses batch endpoints (e.g., OpenAI’s Batch API) to queue multiple prompt requests and process them together for efficiency and cost savings.  
    This approach avoids the script timeout issue and reduces per-token costs. 
    Once the batch results are ready, the backend returns the synthesized data to the add-on, which automatically appends them to a “Generated Data” sheet.

    \item \textbf{Deduplication.}
    To remove near-duplicate data entries, the user can click “Deduplicate” in the Sheets menu. This triggers a Cloudflare Worker that runs the deduplication pipeline (Appendix~\ref{appendix:deduplication_methodology}) and writes the cleaned data to a new sheet. 

    \item \textbf{Data Export.}
    Once satisfied, users can download the curated synthetic dataset (e.g., in CSV or JSON) directly from Google Sheets, integrating seamlessly with broader organizational workflows.
\end{enumerate}

\subsection{Scaling and System Bottlenecks}

While our Google Sheets add-on offers a familiar interface and minimal deployment overhead, several technical constraints can arise at scale:

\begin{itemize}
    \item \textbf{Apps Script Execution Time.}
    Add-ons have a strict 30-second execution limit.
    To mitigate this, we offload heavy tasks (like LLM calls and deduplication) to external backends.
    
    \item \textbf{Cloudflare Workers Resource Limits.}
    The backend runs on Cloudflare Workers, which also has finite CPU and memory quotas (e.g., a few hundred MB per request). 
    To avoid exceeding these limits, we split massive generation jobs into multiple sequential batch requests, rather than submitting a single large payload.
\end{itemize}

\paragraph{Stateless vs.\ Stateful Workflows.}
Each Cloudflare Worker invocation is stateless — no in-memory data persists once a request is completed.
However, the system temporarily retains user-specific data:
the add-on includes a \emph{user\_id} with each request, allowing our backend to associate multi-step operations with the correct spreadsheet session.
Additionally, the deduplication service briefly stores embeddings keyed by \emph{user\_id} (for up to 24 hours) in a database. This limited retention helps streamline the pipeline while maintaining user privacy and conserving resources.
Overall, the design combines the scalability of a stateless, serverless model with just enough short-term state to support multi-step workflows.

\paragraph{Cost Considerations.}
As detailed in Appendix~\ref{appendix:cost_analysis}, LLM usage fees dominate our total costs. However, we observe that OpenAI’s Batch API (or similar endpoints) can reduce per-token costs by up to 50\% compared to synchronous calls. This batch processing approach offers higher rate limits but may delay completion by up to 24~hours, a potential bottleneck for users requiring immediate turnaround. In practice, most batches finish within a few minutes to a few hours, especially for moderate job sizes. Cloudflare Workers itself incurs only modest costs, and Apps Script imposes no additional fees beyond potential usage caps and timeouts.

\section{Case Study}
\label{sec:results}
\begin{table*}[t]
   \centering
   \setlength{\tabcolsep}{5pt}
   \begin{tabular}{l|ccccc|cc}
       \toprule
       \multirow{2}{*}{Model} & \multicolumn{5}{c|}{Performance Metrics} & \multicolumn{2}{c}{Error Rates} \\
       \cmidrule{2-8}
       & Acc. & Brier & Recall & F1 & ROC & False Pos. & False Neg. \\
       \midrule
       Gemma-2b & 0.51 & 0.43 & 0.30 & 0.30 & 0.44 & 0.24 & 0.25 \\
       \quad w/ Real & 0.65 & 0.31 & 0.47 & 0.61 & 0.73 & \underline{0.03} & 0.32 \\
       \quad w/ ELTEX & 0.77 & 0.16 & \underline{0.78} & \underline{0.76} & 0.85 & 0.13 & \underline{0.11} \\
       \quad w/ Real + ELTEX & \underline{0.82} & \underline{0.14} & \textbf{0.79} & \textbf{0.81} & \underline{0.88} & 0.08 & \textbf{0.10} \\
       \midrule
       Llama-Primus & 0.71 & 0.29 & 0.43 & 0.58 & 0.69 & \underline{0.03} & 0.27 \\
       \quad w/ Merged & 0.76 & 0.23 & 0.59 & 0.70 & 0.76 & 0.04 & 0.19 \\
       \midrule
       GPT-4o & \textbf{0.84} & \textbf{0.10} & 0.71 & \textbf{0.81} & \textbf{0.94} & \textbf{0.02} & 0.14 \\
       \bottomrule
   \end{tabular}
   \caption{Performance on social media cyberattack detection. Best scores in \textbf{bold}, second-best \underline{underlined}.}
   \label{tab:performance}
\end{table*}

This section presents our evaluation of ELTEX-generated data for cybersecurity text classification, where each model classifies social media messages for potential relevance for blockchain attacks using a 0-1 scoring system. We focus on practical metrics relevant for industrial applications: classification performance, calibration quality using Brier score \cite{brier1950verification}, and efficiency gains.

Table~\ref{tab:performance} summarizes the performance metrics for all evaluated models: the Gemma-2B base model, fine-tuned variants using LoRA \cite{hu2022lora} (on real data, synthetic data and a hybrid approach) and GPT-4o as a reference point. We include cybersecurity-focused LLMs fine-tuned on the Primus dataset \cite{yu2025primuspioneeringcollectionopensource} to contextualize our results.

\subsection{Task Complexity}
Our setup addresses two key challenges in real-world blockchain security monitoring. First, the model must process multiple messages simultaneously, outputting batch classifications of 10 messages to support high-throughput monitoring scenarios. Second, the model produces calibrated risk scores between 0 and 1, rather than discrete classification, enabling evaluation via Brier score to assess both discrimination and calibration. These scores are returned in structured JSON format (e.g., \texttt{\{"1": 0.9, "2": 0.1\}}).

\subsection{Dataset Construction}
\label{sec:dataset}
Our training pipeline leverages the dataset of 1,603 real-world social media messages, which we first split into training (80\%) and validation (20\%) sets. Using ELTEX, we expanded only the training portion into a synthetic dataset of 8,892 messages (5,530 cyberattack, 3,362 general), with generated classification scores ranging from 0 to 1. 
We collected an independent test set of 398 real-world messages (206 labeled as cyberattack and 192 as general), drawn from entirely distinct attack events not present in our training data. Table~\ref{tab:attack-types} in Appendix \ref{appendix:experimental_details} shows the distribution of primary attack types across training and test sets.

\subsection{Key Findings}

Our evaluation revealed three critical insights for industrial deployment:

\paragraph{Enhanced performance with resource efficiency:} Fine-tuning Gemma-2B with ELTEX-generated data achieved classification performance (F1: 0.81) comparable to GPT-4o while reducing inference costs. This represents significant operational savings for high-volume monitoring scenarios typical in security operations.

\paragraph{Model calibration:} ELTEX-enhanced models demonstrated substantially better calibration (Brier score: 0.14-0.16) than both the base model (0.43) and domain-specialized LLMs, like Primus (0.23-0.29). Well-calibrated confidence scores enable more effective alert prioritization and risk assessment in production environments.

\paragraph{Balanced error profiles:} Our hybrid approach (real data + ELTEX) achieved a favorable balance between false positives (0.08) and false negatives (0.10), unlike models trained solely on real data which showed a strong bias toward one error type (0.03 and 0.32 respectively). This balanced performance enables more reliable threat detection for operational security contexts where both error types carry costs. Notably, the hybrid approach achieved the best overall performance, suggesting complementary benefits from both data sources, which aligns with recent research by \citet{li2024datagenerationusinglarge}.

\section{Conclusion}
ELTEX addresses domain specialization challenges by systematically preserving domain knowledge throughout synthetic data generation. Our cybersecurity case study demonstrated three key advantages: comparable performance to frontier LLMs with significantly reduced computational requirements, superior model calibration, and balanced error profiles critical for operational contexts. The Google Sheets implementation democratizes domain adaptation for non-technical users. Limitations include reliance on seed data for generation and validation limited to a classification task in one domain.

\section*{Ethical Considerations}
\label{sec:ethical}

Our research adheres to strict ethical guidelines to ensure the responsible collection and use of data from social media platforms. The following measures were implemented to address potential ethical concerns:

\paragraph{Compliance with Terms of Service and Privacy Policies} All data collection procedures were conducted in accordance with X’s official API (v2) terms of service and privacy policies \cite{x_api}. We ensured that our data acquisition methods did not violate platform-specific rules or regulations.
    
\paragraph{Data Anonymization} To protect user privacy, we anonymized the collected data by removing all usernames and user mentions. This step ensures that individual users cannot be identified from the dataset.

\paragraph{Google Sheets Implementation}
Our Google Sheets add-on stores user data solely within the user’s spreadsheet, ensuring direct control over content. The add-on minimally processes user email addresses — provided by Apps Script login — for operational purposes (e.g., linking requests to the correct spreadsheet). In line with best practices for personally identifiable information (PII), we neither persist these emails beyond short-term workflows nor share them externally. Users remain responsible for ensuring that any data they enter complies with applicable privacy and ethical guidelines. Meanwhile, embeddings used by the deduplication step are retained externally for up to 24 hours, and then removed. This design respects data ownership and privacy while enabling efficient collaboration and real-time data generation.

\bibliography{anthology,custom}

\clearpage
\appendix
\raggedbottom
\section{ELTEX System Details}
\label{appendix:eltex_system}

\subsection{Overview}
This appendix provides an in-depth overview of the ELTEX methodology used for relevant token extraction from LLMs. The included example aim to enhance the quality, diversity, and relevance of synthetic datasets, particularly within the context of cybersecurity in the blockchain industry. While the examples focus on this domain, the ELTEX approach is adaptable to other domains by modifying task-specific instructions and indicators.

\subsection{ELTEX Prompt Design Workflow}
The ELTEX pipeline integrates the following key steps to maximize the utility of prompts in extracting meaningful tokens and generating high-fidelity synthetic data:

\begin{enumerate}
    \item \textbf{Data Preparation}: Collect real-world data samples and deduplicate them using vector similarity search to ensure uniqueness and variety.
    \item \textbf{Prompt Construction}: Design prompts that combine task descriptions, critical instructions, domain-relevant indicators, and real data samples. 
    \item \textbf{Synthetic Data Generation}: Generate synthetic data using the LLM and apply deduplication to maintain data quality.
\end{enumerate}

\subsubsection{Task Description and Critical Instructions}
A well-crafted task description and critical instructions are essential components of the prompt design that ensure the LLM generates relevant and high-quality synthetic data. These elements provide clear guidance to the model, outlining the objectives and constraints of the data generation task.

\paragraph{Task Description}
The task description defines the primary objective of the synthetic data generation process. It specifies what the model is expected to produce and the context in which the data will be used.

\begin{tcolorbox}[colback=blue!5!white, colframe=blue!75!black, title=Task Description Example]
Your task is to generate a list of social media platform messages to be used as early warning signals for identifying a cyberattack on a blockchain industry participant. Below are two lists. The "Cyberattack Indicators" list contains signals that can lead to financial or reputational loss. The "Social Media Messages" list contains social media platform messages about cyberattacks on a blockchain industry that happened in the past. Use the "Cyberattack Indicators" and "Social Media Messages" to generate 100 new social media platform messages that could imply a cyberattack on a blockchain ecosystem participant, including very early stages of it.
\end{tcolorbox}

\paragraph{Critical Instructions}
Critical instructions provide specific guidelines to ensure the generated data adheres to desired standards and mitigates potential biases.

\begin{tcolorbox}[colback=blue!5!white, colframe=blue!75!black, title=Critical Instructions Example]
Critical:

\begin{itemize}
    \item Use modern vocabulary and writing style.
    \item Leave Law Enforcement and Government Regulators' names unchanged, but replace other named entities (organizations, persons, locations) with fictional, yet plausible and modern ones.
    \item The output must be a list of 100 newly generated social media platform messages without any explanations.
\end{itemize}
\end{tcolorbox}

\paragraph{Implementation Considerations}
When designing task descriptions and critical instructions, consider the following to enhance effectiveness:

\begin{itemize}
    \item \textbf{Clarity and Specificity}: Clearly define the scope and objectives to avoid ambiguous outputs.
    \item \textbf{Bias Mitigation}: Explicitly instruct the model on handling sensitive information and named entities to prevent unintended biases.
    \item \textbf{Adaptability}: Ensure that instructions are adaptable to changes in domain standards or emerging trends.
    \item \textbf{Batch Size Management}: Optimize the number of messages generated per batch to balance quality and computational efficiency.
\end{itemize}

\subsection{Indicators}
This section outlines the process of creating and refining indicators, which are crucial for guiding the LLM in generating relevant synthetic data.

\subsubsection{Indicators Generation}
Indicators serve as a compact prompt section listing concepts relevant to a given use case, fulfilling several crucial purposes:
\begin{itemize}
    \item \textbf{Focuses the LLM's attention}: Concentrates on a highly distilled collection of relevant tokens.
    \item \textbf{Highlights model features}: Emphasizes aspects not well-pronounced in the data samples.
    \item \textbf{Compensates for biases}: Mitigates biases and deficiencies in the data samples.
\end{itemize}

Although a subject-matter expert can manually define this section, relying solely on human expertise may be impractical, as finding domain specialists can be challenging. To mitigate this dependence, we propose leveraging LLMs to generate and refine indicators. By employing the following techniques, we enhance both efficiency and accuracy:

\begin{itemize}
    \item \textbf{Leverage multiple models}: Use LLMs trained on diverse datasets to capture a broad spectrum of perspectives and avoid biases inherent to a single model.
    \item \textbf{Prioritize models with recent knowledge}: Prefer models with the latest knowledge cut-off dates, especially for fast-changing industries and emergent topics.
    \item \textbf{Use different LLM service implementations}: Different implementations might have varying optimizations and safeguards, enhancing the diversity of indicators.
    \item \textbf{Reference public knowledge}: Ground indicators in real-world contexts by incorporating historical events that LLMs are likely familiar with.
    \item \textbf{Use educational resources}: Incorporate resources that would be used to teach manual classification to ensure comprehensive coverage.
\end{itemize}

\begin{tcolorbox}[colback=blue!5!white, colframe=blue!75!black, title=Indicator Generation Prompt Example]
Your task is to generate a list of Blockchain Ecosystem Cyberattack Indicators that could be spotted by looking at social media chatter, including very early stages of the attack. Blockchain ecosystem participants could include centralized and decentralized products, exchanges, protocols, wallets, smart contracts, bridges, oracles, developers, key people, etc. Information included below could help you reason about useful signals for monitoring reports on social media.

General Knowledge:

<article text 1

article text 2

...

article text n>

Historical Events:

<date 1 - named entity 1

date 2 - named entity 2

...

date n - named entity n>
\end{tcolorbox}

\subsubsection{Indicator Summarization}
The outputs from the indicator generation stage often include verbose descriptions with semantic duplicates. It is essential to consolidate and refine the AI-generated content into a concise, non-redundant list of indicators that can be validated by a subject-matter expert. Key points to consider during summarization include:

\begin{itemize}
    \item \textbf{Use a frontier model}: Select models with large context windows, strong reasoning capabilities, and low-temperature settings.
    \item \textbf{Ensure alignment}: Aim for consistency between your understanding of important indicators and the model's output.
    \item \textbf{Avoid manual additions}: Focus on improving the indicator generation prompts rather than manually adding information.
    \item \textbf{Compare multiple models}: Use outputs from different models to ensure no critical information is omitted.
\end{itemize}

\begin{tcolorbox}[colback=blue!5!white, colframe=blue!75!black, title=Indicator Summarization Prompt Example]
Your task is to deduplicate and summarize a list of Blockchain Ecosystem Cyberattack Indicators you will find below. It's okay to merge similar ideas into one concept, but don't remove any ideas completely. Generate a succinct paragraph with densely packed indicators and associated concepts you will find below without additional comments.

<list of indicators>
\end{tcolorbox}

\subsection{Cyberattack Indicators}
This section presents our curated list of cyberattack indicators specific to the blockchain industry. 

\begin{tcolorbox}[colback=blue!5!white, colframe=blue!75!black, title=Cyberattack Indicators List]
Key blockchain ecosystem cyberattack indicators include unusual spikes in transaction volume, abnormal confirmation times, unexpected changes in mining difficulty, and suspicious smart contract interactions. Wallet-related red flags involve sudden activation of dormant addresses, unauthorized transactions, and compromised private keys. Exchange-specific indicators include rapid price fluctuations, withdrawal issues, and unexpected downtime. Infrastructure concerns manifest as unverified nodes, synchronization delays, and potential 51\% attacks. Cross-chain vulnerabilities, security breaches, phishing attempts, and API compromises are critical to monitor. Community sentiment, governance irregularities, and blockchain explorer activity provide additional context for potential threats.
\end{tcolorbox}

\subsection{Handling Alignment and Sensitive Content}
Certain tasks involve sensitive topics, such as cyberattacks, which may trigger alignment safeguards in LLMs, causing them to refuse generating synthetic samples. To mitigate this, the ELTEX system incorporates explicit phrasing in prompt instructions to clarify the context and ethical use of the data. For example, including statements like \textit{``The generated synthetic data will be used solely for research-oriented classification tasks and will not be utilized for any other purposes''} helps in mitigating content generation refusals.

\subsection{Final Prompt Template}
Below is the ELTEX prompt template designed for generating synthetic data.

\begin{tcolorbox}[colback=blue!5!white, colframe=blue!75!black, title=ELTEX Prompt Template for Synthetic Data Generation]
<task description>   

<critical instructions>  

<indicators list>  

<real data samples>   
\end{tcolorbox}

\subsection{ELTEX vs Real Data}
\begin{figure}[t]
    \centering
    \includegraphics[width=0.48\textwidth]{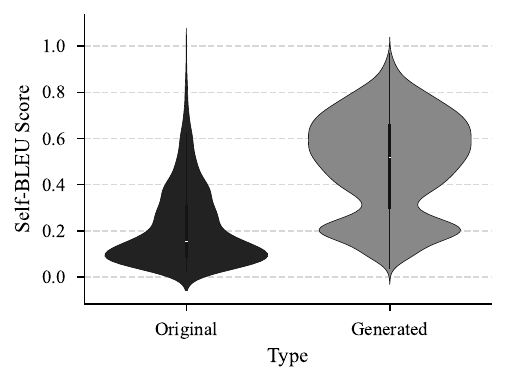}
    \caption{Comparison of Self-BLEU Scores between Generated and Original Data.}
    \label{fig:self-blue}
\end{figure}

The synthetic data exhibits self-BLEU scores ($\mu=0.2831$, $\sigma=0.1791$) comparable to the original messages, demonstrating ELTEX's ability to generate diverse outputs while maintaining domain relevance (see Figure \ref{fig:self-blue}) ~\cite{zhu2018texygen}.

\newpage
\section{Data Annotation with LLMs}
\label{appendix:data_annotation}
\subsection{Prompt Used for Annotation}
The following prompt was used to generate preliminary annotation scores for collected messages. These scores (ranging from 0 to 1) were utilized as a basis for subsequent human review and label refinement.

\begin{tcolorbox}[colback=blue!5!white, colframe=blue!75!black, title=Annotation Prompt]
Your task is to detect early warning signals for cyberattacks on blockchain industry participants. You will be provided with two lists: "Cyberattack Indicators" list contains signals that can lead to a financial or reputational loss. The other list, "Social Media Messages", is an array of objects that should contain messages about cyberattacks on the blockchain industry. Classify those objects by adding `cyberattack\_score` field to the corresponding JSONs and assign a classification score value to it from 0 to 1, where 1 corresponds to an object clearly associated with a potential or existing cyberattack or some context that would imply it, and 0 corresponds to an object that is not connected to a cyberattack at all.

Critical: 

- The output must be only an array from the list "Social Media Messages", only with properties `message\_id` and the corresponding `cyberattack\_score`, machine-readable and formatted as JSON without any explanations, don't add anything else.

- Pay attention to associated timestamps to better understand the context.

- Don't remove any messages.

Cyberattack Indicators

<indicators list>

Social Media Messages:
<real data samples>
\end{tcolorbox}

\subsection{Human Annotation Procedure}
Human annotators used the preliminary labels generated by GPT-4o as a starting point for the annotation process. Guided by the list of cyberattack indicators, they reviewed the assigned cyberattack\_score and adjusted the final labels accordingly. This approach leverages the efficiency of LLM-generated insights while ensuring high annotation quality.

\subsection{Validation and Accuracy}
LLM-annotated labels were additionally validated using a custom function. 

\begin{itemize}
    \item $R = \{r_1, r_2, \dots, r_n\}$: the set of records with true labels ($r_i$), where true labels are those verified by a human annotator.
    \item $A = \{a_1, a_2, \dots, a_m\}$: the set of annotated records with GPT-scores ($a_i$).
    \item $T$: the threshold for classification ($T = 0.5$ by default).
    \item The predicted label $\hat{y}_i$ is defined as:
    \[
    \hat{y}_i = 
    \begin{cases} 
        1, & \text{if } a_i \geq T, \\
        0, & \text{otherwise}.
    \end{cases}
    \]
\end{itemize}

The formula for computing accuracy (\textit{Accuracy}) is:

\[
\text{Accuracy} = 
    \frac{\sum\limits_{i=1}^n \delta(y_i, \hat{y}_i)}{n} \cdot 100
\]

where:
\[
\delta(y_i, \hat{y}_i) = 
\begin{cases} 
    1, & \text{if } y_i = \hat{y}_i, \\
    0, & \text{otherwise}.
\end{cases}
\]

Here, $y_i$ is the true label of record $r_i$, and $\hat{y}_i$ is the predicted label obtained from the annotated record $a_i \in A$ corresponding to $r_i$.

Validation confirmed that GPT-4o achieved over 95\% accuracy.

\newpage
\section{Experimental Setup Details}
\label{appendix:experimental_details}

This appendix provides detailed information about our data, batching strategy, and model configurations. We train three variants of the model: \emph{Real}, \emph{Synthetic}, and \emph{Hybrid}, each with unique data compositions but sharing consistent validation procedures.

\subsection{Batching Strategy and Overall Approach}

All training and inference is conducted in fixed batches of 10 messages. Although this reduces the total “effective” number of parameter-update steps, it aligns with real-world, high-throughput requirements where messages often arrive in bursts. Processing messages in such fixed-size batches simplifies the pipeline and keeps training and inference behavior consistent.

\subsection{Dataset Statistics}
\begin{table*}[t]
\small
\centering
\begin{tabular}{p{4cm}p{5.5cm}p{5.5cm}}
\toprule
\textbf{Primary Attack Type} & \textbf{Training Set (Pre-May 2024)} & \textbf{Test Set (May '24--Jan '25)} \\
\midrule
Social Engineering \& Phishing & Credential theft, wallet phishing & Email compromise, custodian impersonation \\
\midrule
Smart Contract Exploits & Token claim vulnerabilities, flash loans & Token sale exploits, parameter manipulation \\
\midrule
Exchange Security Breaches & Hot wallet compromises, key theft & System access exploits, API vulnerabilities \\
\midrule
DeFi Protocol Attacks & Liquidity pool manipulation, bridge exploits & Cross-chain vulnerabilities, protocol exploits \\
\bottomrule
\end{tabular}
\caption{Distribution of primary attack types across training (1,603 messages) and test (398 messages) sets.}
\label{tab:attack-types}
\end{table*}

We utilize two primary corpora: a \textbf{real-world} dataset and a \textbf{synthetic} dataset derived exclusively from the real training set. The \textbf{hybrid} variant merges these two datasets.  In addition, we collect a \textbf{test set} of entirely new attack events that do not appear in the real dataset, to rigorously assess out-of-distribution performance.

\paragraph{Real Dataset}
The deduplicated real-world dataset contains 1,603 total messages. We split these into training and validation sets using an 80/20 ratio before generating synthetic data:

\begin{itemize}
\item \textbf{Training set (80\%):} 1,283 messages
\begin{itemize}
\item 761 cyberattack-related (59.3\%)
\item 522 general (40.7\%)
\end{itemize}
\item \textbf{Validation set (20\%):} 320 messages
\begin{itemize}
\item 190 cyberattack-related (59.4\%)
\item 130 general (40.6\%)
\end{itemize}
\end{itemize}

After dividing into fixed batches of 10 messages, these total to:
\begin{itemize}
\item \textbf{Training set:} 128 effective batches
\item \textbf{Validation set:} 32 effective batches
\end{itemize}

\paragraph{Synthetic Dataset}
We generated 8,892 synthetic messages (5,530 cyberattack-related and 3,362 general) from the real training set only. This approach avoids data leakage into the validation/test sets. Since we do not create a separate synthetic validation set, the real validation set of 320 messages serves as a consistent benchmark for both synthetic and real training scenarios. The fixed batch size of 10 yields 888 effective training batches in the synthetic-only case.

\paragraph{Hybrid Dataset}
For our hybrid variant, we combine the \emph{entire real training set} (1,283 messages) with \emph{all synthetic messages} (8,892), producing a total of 10,175 training messages. Consequently, the hybrid training set has 1,017 effective batches (10,175 / 10), while the validation set remains the same 320 real messages (32 effective batches). This ensures direct comparability of performance across the real-only, synthetic-only, and hybrid approaches.

\paragraph{Test Set}
We curated an additional real-world test set of 398 messages, consisting of 206 cyberattack-related and 192 general messages. Crucially, these messages come from cyberattack events not present in the original training data, ensuring an independent, out-of-distribution evaluation. This isolated test set allows us to assess how well the model generalizes to novel attack patterns and contexts.

\subsection{Training Configurations}

We tailor batch sizes, epochs, and gradient accumulation to each data scenario.

\subsubsection{Real-Only Model Configuration}
Because the real dataset comprises only 128 effective training batches, we adopt a small per-step batch size and accumulate gradients to maintain stable updates:

\begin{itemize}
\item \textbf{Batch size:} 1
\item \textbf{Gradient accumulation steps:} 16 \quad (\textit{Effective batch size:} $16$)
\item \textbf{Learning rate:} $2\times10^{-4}$
\item \textbf{Max epochs:} 10
\end{itemize}

This setup limits gradient variance and reduces overfitting risks when data is sparse.

\subsubsection{Synthetic-Only and Hybrid Model Configuration}
Both synthetic-only (888 effective training batches) and hybrid (1,017 effective batches) allow more standard batch settings:

\begin{itemize}
\item \textbf{Batch size:} 4
\item \textbf{Gradient accumulation steps:} 8 \quad (\textit{Effective batch size:} $32$)
\item \textbf{Learning rate:} $2\times10^{-4}$
\item \textbf{Max epochs:} 5
\end{itemize}

Since these datasets provide substantially more training examples, a larger effective batch size stabilizes training, and fewer epochs (5) suffice to attain convergence without overfitting.

\subsection{Implementation Details}

All models share the following technical setup:

\begin{itemize}
\item \textbf{Framework:} Hugging Face Transformers \cite{wolf-etal-2020-transformers} and PEFT \cite{mangrulkar2022peft}
\item \textbf{Base model:} Gemma-2B
\item \textbf{Precision:} Mixed precision (\texttt{bfloat16})
\item \textbf{Hardware:} NVIDIA A100 GPU (80GB VRAM)
\item \textbf{LoRA rank:} $r=8$
\item \textbf{Quantization:} 4-bit with \texttt{bfloat16} compute
\item \textbf{Optimizer:} AdamW \cite{loshchilov2018decoupled}
\item \textbf{Input format:} Batches of 10 tokenized messages (Gemma tokenizer)
\item \textbf{Output format:} JSON-structured risk scores
\end{itemize}

Deployments run on Cloudflare Workers AI~\cite{CloudflareWorkersAI2023}, exporting LoRA adapter weights in \texttt{safetensors} format. Inference likewise processes incoming messages in batches of 10, returning per-message risk scores as JSON.

\subsection{Model Selection and Evaluation}

We apply a consistent validation scheme and selection criteria:

\begin{itemize}
\item \textbf{Validation strategy:} Monitoring training/validation losses and Brier score to balance discrimination and calibration
\item \textbf{Early stopping:} Triggered if no metric improvement after 3 consecutive epochs
\item \textbf{Checkpointing:} We store the best model by validation Brier score
\item \textbf{Final evaluation:} Conducted on a separate held-out test set of distinct cybersecurity events
\end{itemize}

By emphasizing the Brier score, we ensure well-calibrated probabilities, crucial in cyberattack scenarios where both false positives and false negatives carry real-world risks.

\newpage
\section{Additional ELTEX Experiments}
\label{appendix:additional_experiments}

This appendix summarizes additional experiments conducted within the ELTEX framework to optimize synthetic data generation. We evaluate key parameters — including deduplication thresholds and temperature settings — and perform clustering analysis to assess semantic diversity and overall data quality for cybersecurity applications.

\subsection{Deduplication Experiments with Real Messages}

\subsubsection{Reduction Statistics}

Deduplication experiments were performed on real messages using similarity thresholds of 0.9 and 0.8. The results are summarized in Table~\ref{tab:dedup_stats_exp}.

\begin{table}[htbp]
    \centering
    \renewcommand{\arraystretch}{1.2} 
    \small 
 \begin{tabular}{|>{\centering\arraybackslash}l|c|c|c|}
        \hline
        \textbf{Type} & \textbf{Threshold} & \textbf{Original} & \textbf{Reduced} \\ \hline
        Cyberattack    & 0.9             & 1,078          & 951           \\ \hline
        Cyberattack    & 0.8             & 1,078          & 413           \\ \hline
        General        & 0.9             & 688            & 652           \\ \hline
        General        & 0.8             & 688            & 446           \\ \hline
    \end{tabular}
    \caption{Reduction Statistics for Cyberattack and General Messages}
    \label{tab:dedup_stats_exp}
\end{table}

\subsubsection{Clustering Analysis on Real Data}

Clustering analysis was conducted using DBSCAN to evaluate the semantic coherence of the deduplicated datasets at thresholds of 0.9 and 0.8. The analysis revealed key differences in the impact of each threshold:

\begin{itemize}
    \item \textbf{Threshold 0.8:} This more aggressive threshold removed a substantial number of messages, resulting in datasets with less semantic overlap but also reduced topical coverage. Messages retained under this threshold often lacked sufficient context for meaningful clustering, as many subtle variations were eliminated during the deduplication process.
    \item \textbf{Threshold 0.9:} A more moderate approach, this threshold preserved a higher proportion of messages, maintaining greater semantic diversity. The retained dataset better represented nuanced variations in language use, enabling more coherent clustering into distinct topical groups. This balance allowed for the identification of meaningful patterns within both cyberattack-related and general messages.
\end{itemize}

\subsection{Temperature Experiments for Synthetic Data Generation}

\subsubsection{Retention Trends}

\begin{figure}[htbp]
    \centering
    \begin{subfigure}{0.45\textwidth}
        \centering
        \includegraphics[width=\textwidth]{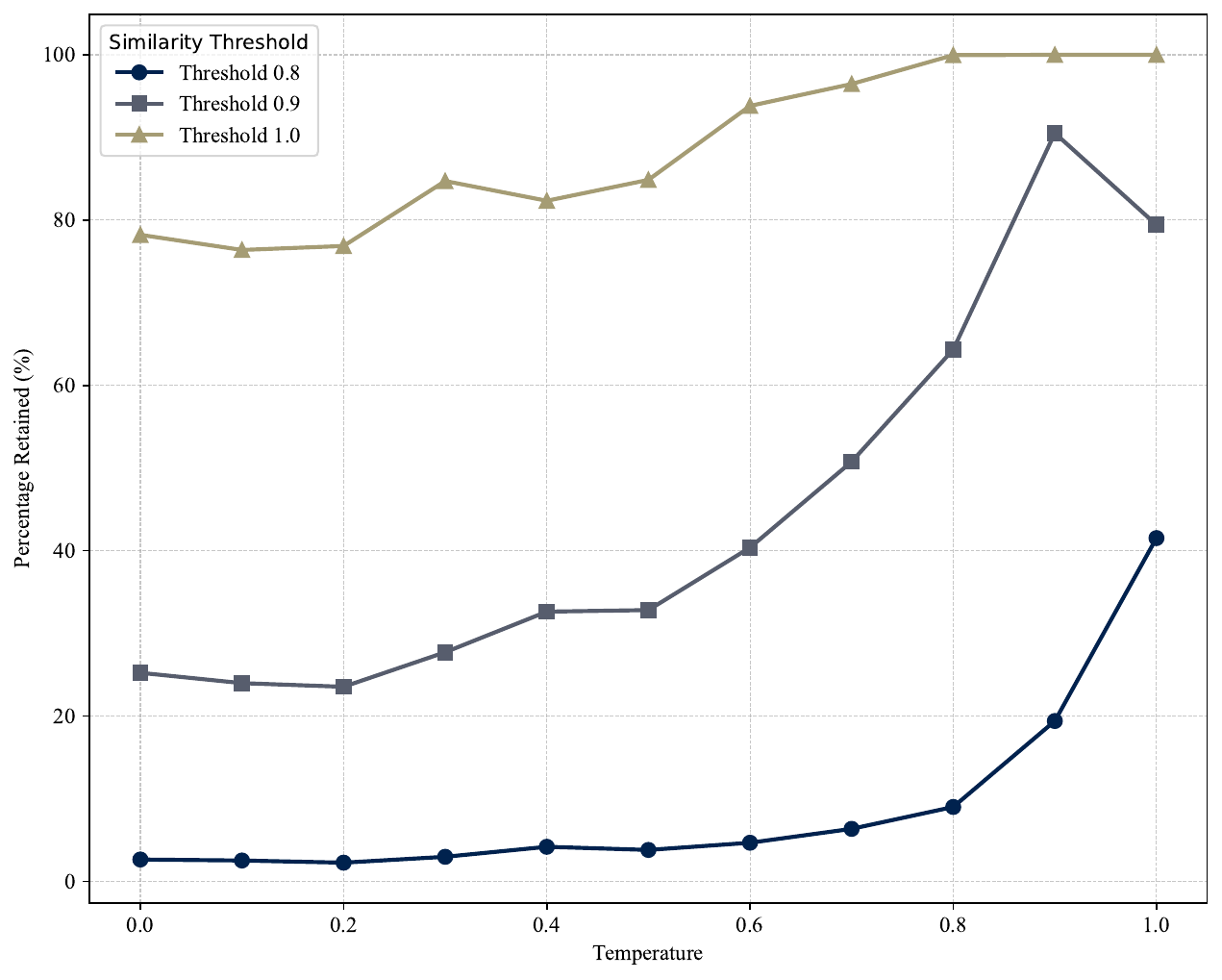}
        \caption{Cyberattack Messages}
        \label{fig:retention_cyberattack}
    \end{subfigure}
    \hfill
    \begin{subfigure}{0.45\textwidth}
        \centering
        \includegraphics[width=\textwidth]{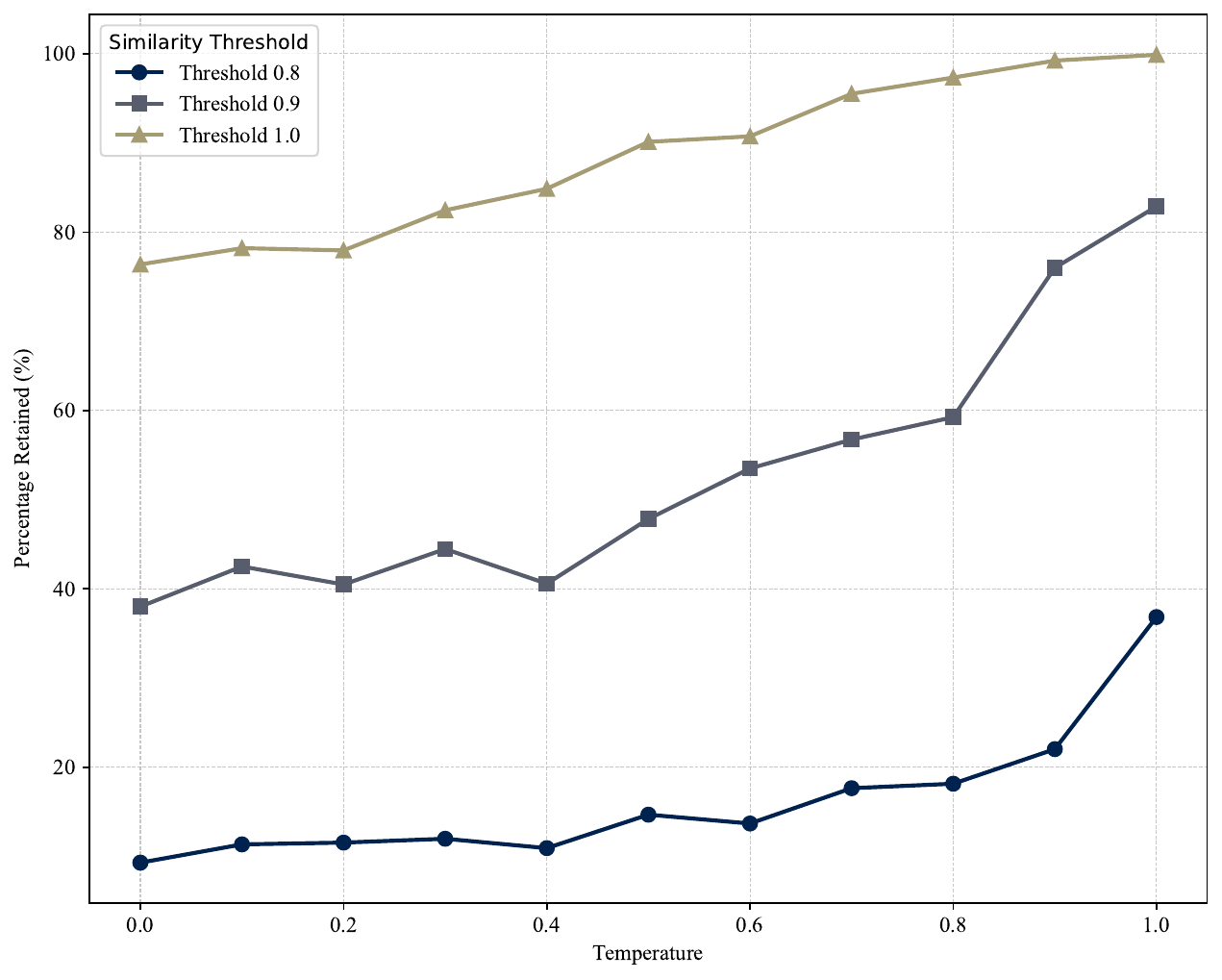}
        \caption{General Messages}
        \label{fig:retention_general}
    \end{subfigure}
    \caption{Retention percentage across different temperature settings and similarity thresholds (0.8, 0.9, 1.0). Note that even with threshold 1.0, exact duplicates are still removed by the deduplication service.}
    \label{fig:retention_temp}
\end{figure}

Synthetic data generation experiments assessed the impact of temperature settings (0.0 to 1.0) on data retention. Retention percentages for different temperatures are illustrated in Figure~\ref{fig:retention_temp}.

\paragraph{Key Observations}
\begin{itemize}
    \item \textbf{Higher Temperatures (0.7–1.0):} These settings produced more diverse outputs, resulting in higher retention rates.
    \item \textbf{Lower Temperatures (0.0–0.6):} These settings generated repetitive messages, leading to lower retention rates. 
    \item \textbf{Optimal Balance:} A temperature of 0.8 provided a balance, yielding both high retention rates and semantically diverse messages without sacrificing coherence.
\end{itemize}

\paragraph{Interpretation and Insights}
The experiments suggest that temperature settings have a moderate influence on the diversity of synthetic data. Higher temperatures tend to introduce more variation in the generated messages, while lower temperatures result in more consistent but potentially repetitive outputs. For cybersecurity datasets, a temperature of 0.8 struck a practical balance, maintaining sufficient variation in generated messages.

\paragraph{Note on Generation Variability}
Due to the stochastic nature of LLM-based generation and our prompt construction methodology, we observed some variation in retention rates across different generation runs at the same temperature setting. Following our approach, real messages were randomly shuffled before being grouped into sets of 10 examples per prompt, resulting in slightly different prompt compositions across generation runs. This, combined with the inherent randomness in the generation process, led to variations in retention rates. For instance, at temperature 0.8 with a similarity threshold of 0.9, retention rates varied between 60-70\% across different generation sessions. This variability is expected and reflects both the randomized prompt construction and the stochastic nature of LLM outputs. Despite these fluctuations, the relative trends and optimal temperature recommendations remain consistent.

\newpage
\section{Deduplication Methodology}
\label{appendix:deduplication_methodology}

\subsection{Deduplication Process Overview}
To ensure the quality and diversity of synthetic data, we implemented a two-stage deduplication pipeline that combines exact match filtering with semantic similarity analysis. This methodology effectively eliminates duplicate or near-duplicate messages, maintaining the integrity and variability of the dataset.

\subsubsection{Initial Deduplication}
The first stage implements the exact content matching using set operations to eliminate identical entries. Given a batch of messages $M = \{m_1, m_2, \ldots, m_n\}$, we construct a set $S$ of unique content:
\begin{equation}
   S = \{content(m) \mid m \in M\}
\end{equation}
where $content(m)$ represents the textual content of message $m$. This operation reduces the computational overhead for subsequent semantic analysis by eliminating exact duplicates early in the pipeline.

\subsubsection{Embedding Similarity Analysis}
The second stage employs embeddings to identify and filter semantically similar content. For each unique message, we generate vector representations using the BAAI BGE base model (BGE-base-en-v1.5) \cite{bge_embedding} deployed through Cloudflare Workers AI \cite{CloudflareWorkersAI2023}. Given a message $m$, its embedding $e(m) \in \mathbb{R}^d$ is computed through the model:

\begin{equation}
   e(m) = \text{BGE}(content(m))
\end{equation}

where $d$ represents the embedding dimension. To determine similarity between messages, we employ cosine distance:

\begin{equation}
   \scalebox{0.8}{$similarity(m_1, m_2) = 1 - cosine\_distance(e(m_1), e(m_2))$}
\end{equation}

Messages are processed in batches of size $k$ (where $k = 100$ in our implementation) to optimize computational efficiency while maintaining reasonable memory usage. For each new message $m$, we compute its maximum similarity score against existing entries in the database $D$:

\begin{equation}
   \scalebox{0.8}{$max\_similarity(m) = \max\{similarity(m, d) \mid d \in D\}$}
\end{equation}

A message is considered sufficiently unique and retained if its maximum similarity score falls below a configurable threshold $\varepsilon$:

\begin{equation}
   is\_unique(m) \iff max\_similarity(m) < \varepsilon
\end{equation}

\subsection{Implementation Details}
The deduplication process is implemented using scalable and efficient algorithms to handle large volumes of data. Specific implementation details, including parameter settings and optimization techniques, are provided below:

\begin{itemize}
    \item \textbf{Similarity Threshold ($\varepsilon$):} We set the threshold $\varepsilon = 0.9$ to balance between removing duplicates and preserving unique variations. Experiments with alternative thresholds (e.g., 0.8) are detailed in \textbf{Appendix C}.
    \item \textbf{Batch Processing:} Messages are processed in batches to enhance computational efficiency and manage memory usage effectively.
    \item \textbf{Embedding Generation:} The BAAI BGE base model is chosen for its robustness in generating high-quality embeddings suitable for semantic similarity tasks.
    \item \textbf{Integration with Database:} The deduplication system is integrated with our database to maintain a clean and diverse dataset, facilitating accurate model training and evaluation.
\end{itemize}

\subsection{Deduplication Service Features}
The system maintains separate tables for different data categories (production, research, and scraped data) while preserving metadata such as:
\begin{itemize}
    \item \textbf{Topic Classification:} Assigning messages to predefined topics.
    \item \textbf{Industry Categorization:} Grouping messages based on industry relevance.
    \item \textbf{Timestamp Information:} Ensuring chronological consistency.
    \item \textbf{Source Metadata:} Capturing platform-specific attributes.
    \item \textbf{Message Identifiers:} Ensuring uniqueness using hashed IDs.
\end{itemize}

\subsection{Monitoring and Performance Metrics}
To ensure real-time quality control, the deduplication pipeline tracks the following metrics:
\begin{enumerate}
    \item \textbf{Total Messages Received:} Number of messages processed.
    \item \textbf{Messages Retained:} Count of messages passing deduplication.
    \item \textbf{Messages Filtered:} Number of duplicates removed.
    \item \textbf{Insertion Rate:} Ratio of retained to received messages.
\end{enumerate}

These metrics enable continuous monitoring of data quality and diversity throughout the collection process.

\newpage
\section{Costs Analysis}
\label{appendix:cost_analysis}

Our synthetic data generation relied on the GPT-4o model served via Azure OpenAI. The pricing structure was as follows:
\begin{itemize}
    \item Input tokens: \$2.50 per million tokens
    \item Output tokens: \$10.00 per million tokens
\end{itemize}

\paragraph{Request Configuration:}
Each request consisted of:
\begin{itemize}
    \item A shared extraction prompt containing up to ten real examples
    \item Generation settings configured to output 100 synthetic messages
\end{itemize}

\paragraph{Token Usage:}
Per single request:
\begin{itemize}
    \item Input tokens: approximately 500-800 tokens
    \item Output tokens: approximately 1,500-2,100 tokens
\end{itemize}

\paragraph{Cost Breakdown:}
\begin{itemize}
    \item Cost per request: approximately \$0.016-\$0.023
    \item Cost per synthetic message: approximately \$0.00016-\$0.00023
    \item Total cost for 16,030 samples: approximately \$3-\$4
\end{itemize}

\paragraph{Additional Costs:}
Message embedding using the \texttt{cf/baai/bge-base-en-v1.5} model (\$0.008 per million tokens) added negligible costs (<\$1) to the total budget, as each message required only a few dozen tokens for embedding.
\end{document}